# Development and Validation of a Machine Learning Algorithm for Clinical Wellness Visit Classification in Cats and Dogs

Donald Szlosek[1*], Michael Coyne[1], Julia Riggot[1], Kevin Knight[1], DJ McCrann[1], Dave Kincaid[1]

[1] *IDEXX Veterinary Laboratories, Inc., 1 IDEXX Drive, Westbrook, ME 04092 USA*

*Corresponding Author

## Introduction

Early disease detection in veterinary care relies on identifying subclinical abnormalities in asymptomatic animals during wellness visits. This study introduces an algorithm designed to distinguish between wellness and other veterinary visits.

## Objectives:

The purpose of this study is to validate the use of a visit classification algorithm compared to manual classification of veterinary visits by three board-certified veterinarians.

## Materials and Methods

Using a dataset of 11,105 clinical visits from 2012 to 2017 involving 655 animals (85.3% canines and 14.7% felines) across 544 U.S. veterinary establishments, the model was trained using a Gradient Boosting Machine model. Three validators were tasked with classifying 400 visits, including both wellness and other types of visits, selected randomly from the same database used for initial algorithm training, aiming to maintain consistency and relevance between the training and application phases; visit classifications were subsequently categorized into "wellness" or "other" based on majority consensus among validators to assess the algorithm's performance in identifying wellness visits.

## Results

The algorithm demonstrated a specificity of 0.94 (95% CI: 0.91 to 0.96), implying its accuracy in distinguishing non-wellness visits. The algorithm had a sensitivity of 0.86 (95% CI: 0.80 to 0.92), indicating its ability to correctly identify wellness visits as compared to the annotations provided by veterinary experts. The balanced accuracy, calculated as 0.90 (95% CI: 0.87 to 0.93), further confirms the algorithm's overall effectiveness.

**Clinical Significance**

The algorithm exhibits strong specificity and sensitivity, ensuring accurate identification of a high proportion of wellness visits. Overall, this algorithm holds promise for advancing research on preventive care's role in subclinical disease identification, but prospective studies are needed for validation.

**Introduction**

Medical records can provide a wealth of information to support studies of disease prevalence. Within the human medical field, standardized medical coding for insurance and the International Coding of Disease (ICD-11) can facilitate classification of visit types or diagnoses across clinics and hospitals[1]. Veterinary medicine lacks a similar classification system that allows for standardized recognition of clinical behavior at visits, and this has been a barrier to studies investigating clinical behavior, results, or outcomes in the field[2].

Wellness or preventive care visits play an important role in the veterinary-client-patient relationship and provide an opportunity to educate clients and identify subclinical disease[3,4]. There can be substantial variation between clinics in what is included within a wellness visit and there is no general agreement on what defines a wellness visit or what should be included in wellness care[4]. Depending on the clinic, a wellness visit could be defined only as the examination of apparently healthy animals with no health concerns, to visits that include preventive care and routine laboratory testing of blood, feces and urine in addition to the physical exam. There is limited information on the results of routine bloodwork—meaning tests ordered without intending diagnosis or monitoring—at various life stages. Most studies that have explored laboratory results in healthy dogs and cats typically have a narrow scope[5–8]. They might concentrate on one specific breed, consider only a restricted range of analytes, focus primarily on older pets, or implement a strict definition of what defines "healthy" pets. Currently, there are limited large scale real-world evidence (RWE) studies on the value of wellness visits. A major barrier in the collection of large-scale RWE studies is the inability to determine if a clinical examination constitutes a wellness visit due to the non-uniform collection of clinical information.

This study aims to validate a machine learning algorithm designed to classify wellness visits for dogs and cats that have been presented to veterinary practices across the United States. The performance of the algorithm will be benchmarked as compared to the consensus results of three licensed veterinarians who classified a clinical examination as a wellness visit or other visit.

**Algorithm Training**

The purpose of the visit classification algorithm is to use electronic practice information management records to determine why a pet owner visited their veterinarian (i.e. Why did the pet owner bring their animal to the practice?). A patient visit was defined as when a patient (or the owner of the patient) of a clinic comes to the clinic and obtains one or more products or services on behalf of an animal. The visit begins when the animal walks through the clinic door and ends when the animal leaves the clinic. A patient visit may also be when an animal owner comes to the clinic and obtains some product on behalf of the animal. Each visit was reviewed and classified based on what the intent of the visit was by the pet owner. Intent was determined by several factors available in the database including: the time since the last visit, appointment notes, invoice items or medical notes if available. Labels were focused on the intent of the pet owner for the visit. If the intent was unclear based on information provided, or if the record was an administrative line item and not something that represents activity for the pet, it was excluded. During the initial human annotation process, visits were first classified into clinical or non-clinical visit categories. Visits were then classified further as wellness, non-wellness, and non-clinical visits (consisting of boarding, grooming, and retail). Annotators were masked to visit classifications applied by other annotators. Visits were organized by patient at a single clinic. A custom annotation tool was developed to anonymously monitor annotator consensus. It tracked individual annotator metrics, training set agreement, and group agreement on new labels (blind to individual annotators).

A random selection of 11,105 clinical visits from 2012 to 2017 was used to train our model. These visits involved 655 animals (85.3% canines and 14.7% felines) from 544 veterinary establishments in the United States. The median duration of a visit was one day (IQR: 1.0 – 1.0 days) and the median number of transaction line items per visit was two transactions (IQR: 1 - 5 transactions). We classified the visits into four categories: wellness (24.5%), sick (23.1%), non-clinical (42.5%), and unknown (10.0%). Non-clinical visits included boarding, grooming, and retail services. A single visit could have multiple labels with the exception of a retail visit which was defined to be mutually exclusive to all the other categories.

During the algorithm development, to annotate the visits for intent, we used two methods. First, we asked a pair of veterinarians to label each of the 5,984 visits in the preliminary phase (Supplemental Figure 1A). They agreed on 5,058 visits and labeled the remaining 926 as unknown. These unknown visits were excluded from the initial model training. Second, we asked one of six board-certified veterinarians to label each of the additional 5,121 visits. These visits were added to the training dataset along with the agreed-upon 5,058 visits from the preliminary phase.

A comprehensive set of features was used to classify visits accurately using a Gradient Boosting Machine (GBM) model using the H20 3.20.0.2 and R version 3.5.1[9,10]. These features encompassed various demographic, clinical and transactional aspects of the visits. By leveraging this diverse array of features, the algorithm aimed to capture nuanced patterns and relationships in the data, ultimately enhancing its ability to distinguish between wellness visits and other visits in veterinary clinics. The classification model for the classification of wellness visits, was trained employing a 5-fold cross-validation approach. This process resulted in performance metrics including an F1 score of 0.93, a recall of 0.93, a precision of 0.93, and a specificity of 0.97 for the detection of wellness visits as measured on a held-out test dataset.

**Training validation annotators**

The three veterinarian validators were put through an education period using 125 visits (25 from each of the five classes - as wellness, non-wellness, boarding, grooming, and retail) to become familiar with the annotation tool (Supplemental Figure 1B). These visits were selected from the initial 5,058 visits where the label was agreed upon by two veterinarians and were used to train the algorithm (see Algorithm Training Section above). The veterinarians were granted access to the labels that had been collaboratively determined by the two initial veterinarians who categorized the visit. As part of the education process, round table virtual discussions were allowed to gain alignment on classification of visits. These discussions were facilitated by an expert veterinarian involved in the initial training of the algorithm (MC) and the data scientists involved in the development of the algorithm (JR, DM).

After the education period, each of the validators were assessed for agreement to a random selection of 100 visits that matched the distribution of visit types seen in production. The veterinarians were blinded to the label of the visits for this assessment. These visits were selected from the initial 5,058 visits where the label was agreed upon by two veterinarians and were used to train the algorithm (see Algorithm Training Section above). A benchmark of 85% agreement (defined *pre-hoc*) to the initial agreed upon label by the two veterinarians used to develop the training dataset was required for the validators to continue to the validation study.

**Development of validation dataset**

After the educational period, the three validators were assigned the task of classifying the same 400 visits, distinguishing between wellness visits and other types of visits. To ensure that these 400 visits reflected the distribution of the population of visits expected in a live environment, they were randomly selected from the same database that was used for the initial training of the algorithm. The selected visits were not part of the data used to train the algorithm. This was done to ensure consistency and relevance between the training and application phases. To assess the performance of the algorithm for the identification of wellness visits, the results were split into two categories: 1) "wellness" containing the wellness visit type; 2) "other" containing all visits not defined as a wellness visit (for example non-wellness and non-clinical visits). The reference label for each visit was then generated by using the majority consensus for "wellness" or "other" by each of the validators.

**Statistical Analysis**

A sample size of 400 visits was considered adequate for the identification of wellness visits based off bootstrap simulation using one of the holdout datasets from the k-fold cross validation from the training dataset. In the case that the validators finished their caseload with additional time, they were allowed to continue annotating visits until budget/ time ran out. As a result, a total of 800 cases (double the required amount based on sample size calculation) were collected to accommodate any extra time validators had for labeling additional visits. Validators were permitted to continue annotating until the budget was exhausted. Consequently, the final validation dataset comprised 622 visits. Initial assessment of validator performance against each other was compared using percent exact match across all three raters and Fleiss Kappa for agreement. The performance of the algorithm in accurately classifying wellness and other visits was evaluated using several metrics, including Sensitivity (Recall), Specificity, Positive Predictive Value (PPV, Precision), Negative Predictive Value (NPV), F1 Score, Balanced Accuracy, Matthews Correlation Coefficient (MCC), and Jaccard Index. To get additional measure of model performance the conditional independence model using the Expectation-Maximization (EM) algorithm to estimate disease prevalence as well as the sensitivity and specificity of diagnostic tests in the absence of a perfect gold standard [11,12]. To determine the confidence intervals and estimates for these metrics, a bootstrapping approach was applied. Specifically, 2,000 bootstrap samples were generated by randomly sampling visits with replacement from the original dataset. Bootstrap validation and calibration plots were performed using the Hmisc and rms packages [13,14]. Statistical analysis was done using R version 4.0.2 and various helper functions from the tidyverse[9,15]. Data visualization was generated using ggplot2[16].

**Results**

A total of 622 total visits were classified by all three trained validating veterinarians. Most were dog visits (78.2%, n = 487) with 135 cat visits (21.7%). The median age of dogs was 6.0 years (IQR: 2.0 - 9.3 years) and of cats was 6.0 years (1.7 - 12.0) with a single cat with no recorded age. A full list of demographic information can be found in Table 1.

Table 1. Pet demographic information from 622 veterinary visits.

|  | Dog Visits | | Cat Visits | |
| --- | --- | --- | --- | --- |
|  | n | % | n | % |
| **Sex** | | | | |
| Female | 33 | 6.8 | 6 | 4.4 |
| Female Spayed | 200 | 41.1 | 55 | 40.7 |
| Male | 44 | 9 | 6 | 4.4 |
| Male Neutered | 209 | 42.9 | 67 | 49.6 |
| Unknown | 1 | 0.2 | 1 | 0.7 |
| **Life Stage** | | | | |
| Juvenile | 90 | 18.5 | 23 | 17 |
| Young adult | 92 | 18.9 | 13 | 9.6 |
| Mature adult | 120 | 24.6 | 50 | 37 |
| Senior | 85 | 17.5 | 34 | 25.2 |
| Geriatric | 100 | 20.5 | 14 | 10.4 |
| Unknown | --- | --- | 1 | 0.7 |

Cat Life Stage: Kitten (<= 1 year), Young Adult (>1 to <= 2 years), Mature adult (> 2 years <= 10 years), senior (> 10 years to <= 15 years), geriatric (> 15 years).
Canine Life Stage: Puppy (<= 1 year), Young Adult (>1 to <= 4 years), Mature adult (> 4 years <= 7 years), senior (> 7 years to <= 10 years), geriatric (> 10 years).

**Interobserver Agreement**

Of the 622 visits, all three trained validators agreed on the classification of 96.9% (602) of the visits with 457 (73.5%) visits being classified as other and 146 (23.5%) being classified as wellness (Fleiss Kappa = 0.91, Supplemental Table 1, Supplemental Appendix 1). The proportion of visits that have a reference label of wellness is 23.3% (n = 145), with 24.8% (n = 121) wellness visits classified for dogs and 17.8% (n = 24) classified as wellness for cats.

**Supplemental Table 1.** Classification of Visits by Validators with Reference Labels in cats and dogs.

| Validator 1 | Validator 2 | Validator 3 | n | % | Reference Label |
|---|---|---|---|---|---|
| Other | Other | Other | 457 | 71.9 | Other |
| Other | Other | Wellness | 10 | 1.6 | Other |
| Other | Wellness | Other | 1 | 0.2 | Other |
| Other | Wellness | Wellness | 6 | 0.9 | Wellness |
| Wellness | Other | Other | 5 | 0.8 | Other |
| Wellness | Other | Wellness | 5 | 0.8 | Wellness |
| Wellness | Wellness | Other | 6 | 0.9 | Wellness |
| Wellness | Wellness | Wellness | 146 | 23 | Wellness |

**Algorithm Performance**

The algorithm demonstrated a specificity of 0.94 (95% CI: 0.91 to 0.96), implying its accuracy in distinguishing non-wellness visits (Table 2-3). The algorithm had a sensitivity of 0.86 (95% CI: 0.80 to 0.92), indicating its ability to correctly identify wellness visits as compared to the annotations provided by veterinary experts (Table 2-3). This suggests that the algorithm accurately identified 86% of wellness visits. These results indicate that the algorithm accurately recognized 94% of non-wellness and had a low false positive rate. The balanced accuracy, calculated as 0.90 (95% CI: 0.87 to 0.93), further confirms the algorithm's overall effectiveness (Table 3). In addition, the model was observed to be well calibrated across the predictive range (Supplemental Figure 2). Employing the conditional independence model with the EM algorithm to account for an imperfect gold standard, we obtained a specificity of 0.97 (95% CI: 0.95 – 0.99) and a sensitivity of 0.85 (95% CI: 0.79 – 0.90). Additional measures of model performance are found in Table 3 and in Supplemental Appendix 2.

**Table 2.** Contingency table of algorithm performance to correctly identify wellness visits as compared to reference method (majority of three veterinarians).

|  |  | Reference Method (3 Annotators) |  |  |
| --- | --- | --- | --- | --- |
|  |  | Wellness | Other | Total |
| **Algorithm** | Wellness | 20.1% | 5.0% | 25.1% |

|  | (125) | (31) | (156) |
|---|---|---|---|
| Other | 3.2% | 71.7% | 74.9% |
|  | (20) | (446) | (466) |
| Total | 23.3% | 76.7% | 100.0% |
|  | (145) | (477) | (622) |
|  | Sensitivity | Specificity |  |
|  | 86.2% | 93.5% |  |

**Table 3.** Bootstrapped model performance of algorithm and 95% confidence interval estimates

| Performance Metrics | Estimate | Lower CI | Upper CI |
|---|---|---|---|
| Sensitivity (Recall) | 0.86 | 0.80 | 0.92 |
| Specificity | 0.94 | 0.91 | 0.96 |
| PPV (Precision) | 0.80 | 0.74 | 0.86 |
| NPV | 0.96 | 0.94 | 0.97 |
| F1 Score | 0.83 | 0.78 | 0.87 |
| Balanced Accuracy | 0.90 | 0.87 | 0.93 |
| MCC | 0.78 | 0.72 | 0.83 |
| Jaccard Index | 0.71 | 0.64 | 0.78 |

F1 represents the harmonic mean of recall and precision. PPV = Positive Predictive Value (Precision), NPV = Negative Predictive Value, MCC = Mathew's Correlation Coefficient.

**Discussion**

Early disease detection in veterinary care depends on the identification of subclinical abnormalities in asymptomatic animals which could be evaluated during a wellness visit[5]. Providing a method to determine the type of veterinary visit is essential to determining the benefit of wellness visits. The algorithm developed for this study demonstrated strong specificity and sensitivity, suggesting a robust ability to distinguish between wellness and other visits. The high specificity and sensitivity confirm that the algorithm was able to accurately identify a high proportion of wellness visits. A high specificity in our classification model minimizes the risks associated with incorrectly categorizing other visits as wellness visits. Such a misclassification would have more serious consequences compared to inadvertently classifying a wellness visit as a non-wellness one, as it could potentially lead to overlooking individuals who are in actual need of medical attention or intervention.

This study has limitations that warrant consideration. Primarily, the identification and classification of visit types were conducted using data from specific practice information management systems, which may limit the algorithm's generalizability across different clinical settings and necessitates additional validation. Moreover, as visit type evaluation is not a typical part of the regular veterinary clinical workflow, direct comparisons of the algorithm's ability to identify wellness visits to a single veterinarian could be challenging to interpret. The model performance was dependent on the annotations provided by veterinary experts. Our interobserver agreement results, with a Fleiss Kappa of 0.91, underline the consistency among expert annotators, providing a reliable basis for the algorithm's training. The reliance of the algorithm's training on the agreement among veterinarians in classifying visits underscores the pivotal role of expert judgment in this process. It also brings to light the limitations of the algorithm in situations where there is a lack of consensus among experts. To ensure practical relevance, future efforts should aim to integrate this aspect into routine workflow and use more varied data for algorithm training. Additionally, our sample was skewed towards dogs (78.2% of visits), which may have influenced the algorithm's performance across species. Future research should investigate further refining the algorithm's ability to differentiate between wellness and sick visits, especially in cases where expert consensus might be challenging. In particular, the discrepancy in wellness visit classification between cats and dogs calls for a more nuanced approach to species-specific care patterns in future algorithm development.

Wellness and preventive care visits are crucial in the veterinary-client-patient relationship, offering a chance to inform clients and detect underlying diseases early. The classification of wellness visits in large-scale RWE data could expand research on the role preventive care plays in the identification of subclinical disease. While the utilization of a visit classification algorithm on retrospective RWE data may expedite the assessment of the clinical value of preventive care visits, further prospective studies are essential to validate these findings.

**References**

1. ICD-11. https://icd.who.int/en.
2. Lustgarten, J. L., Zehnder, A., Shipman, W., Gancher, E. & Webb, T. L. Veterinary informatics: forging the future between veterinary medicine, human medicine, and One Health

initiatives—a joint paper by the Association for Veterinary Informatics (AVI) and the CTSA One Health Alliance (COHA). *JAMIA Open* **3**, 306–317 (2020).

3. American Animal Hospital Association. Preventive Care Pet Health Resources | AAHA. https://www.aaha.org/practice-resources/pet-health-resources/preventive-care/.

4. Janke, N., Coe, J. B., Bernardo, T. M., Dewey, C. E. & Stone, E. A. Use of health parameter trends to communicate pet health information in companion animal practice: A mixed methods analysis. *Veterinary Record* **190**, e1378 (2022).

5. Willems, A. *et al.* Results of Screening of Apparently Healthy Senior and Geriatric Dogs. *J Vet Intern Med* **31**, 81–92 (2017).

6. Dell'Osa, D. & Jaensch, S. Prevalence of clinicopathological changes in healthy middle-aged dogs and cats presenting to veterinary practices for routine procedures. *Aust Vet J* **94**, 317–323 (2016).

7. Paepe, D. *et al.* Routine health screening: findings in apparently healthy middle-aged and old cats. *J Feline Med Surg* **15**, 8–19 (2013).

8. Jeffery, U., Jeffery, N. D., Creevy, K. E., Page, R. & Simpson, M. J. Variation in biochemistry test results between annual wellness visits in apparently healthy Golden Retrievers. *J Vet Intern Med* **35**, 912–924 (2021).

9. R Development Core Team. *R: A Language and Environment for Statistical Computing*. (R Foundation for Statistical Computing, 2009).

10. Fryda, T. *et al.* h2o: R Interface for the 'H2O' Scalable Machine Learning Platform. (2023).

11. Walter, S. D. & Irwig, L. M. Estimation of test error rates, disease prevalence and relative risk from misclassified data: a review. *J Clin Epidemiol* **41**, 923–937 (1988).

12. Dawid, A. P. & Skene, A. M. Maximum Likelihood Estimation of Observer Error-Rates Using the EM Algorithm. *Journal of the Royal Statistical Society. Series C (Applied Statistics)* **28**, 20–28 (1979).

13. Harrell, F. Hmisc: Harrell Miscellaneous. (2021).

14. Harrell, F. rms: Regression Modeling Strategies. (2021).

15. Wickham, H. & RStudio. tidyverse: Easily Install and Load the 'Tidyverse'. (2017).

16. Wickham, H. *et al.* ggplot2: Create Elegant Data Visualisations Using the Grammar of Graphics. (2018).

**Supplemental Data**

**Supplemental Appendix 1.** Initial Annotator Training Performance

In this study, the performance of three annotators, referred to as Validator 1, Validator 2, and Validator 3, in labeling wellness and other visits were evaluated. The annotators' match rates were calculated by comparing their annotations to the reference labels that were annotated to train the visit classification algorithm. Validator 1 achieved a match rate of 97.3%, with two instances where visits labeled as "other" were classified as "wellness" according to the reference labels. Validator 2 achieved a match rate of 96.0%, with two "other" visits misclassified as "wellness" and one "wellness visit" misclassified as "other." Validator 3 achieved a match rate of 95.8%, with three "other" visits mistakenly labeled as "wellness." These findings provide insights into the annotators' accuracy and allowed to confidence in moving forward with the validation of the algorithm.

**Supplemental Appendix 2.** Model Calibration and additional performance metrics.

In our analysis, we used several statistical measures to evaluate prediction accuracy. The Brier Score was 0.074, indicating good probabilistic prediction accuracy. The Harrel's C Index was 0.879, showing strong model discrimination. Calibration curve metrics revealed an index-corrected Emax of 0.008, an index-corrected slope of 0.9686, and an intercept of 0.0073. The near-zero intercept suggests minimal constant or proportion bias across the prediction domain. Overall, these metrics confirm the model's reliability and prediction accuracy.

**Supplemental Figure 1.** **A)** Two methods were used annotate visits for intent during algorithm development. First, two veterinarians labeled 5,984 visits, agreeing on 5,058, with the remaining 926 marked as unknown and excluded from initial training (Preliminary Phase). Second, a board-certified veterinarian labeled an additional 5,121 visits, which were added to the training dataset. **B)** Veterinarian validators underwent education using 125 visits to familiarize themselves with the annotation tool and participated in virtual discussions for alignment. Afterward, validators were assessed for agreement on a random selection of 100 visits, requiring an 85% agreement benchmark to proceed to the validation study.

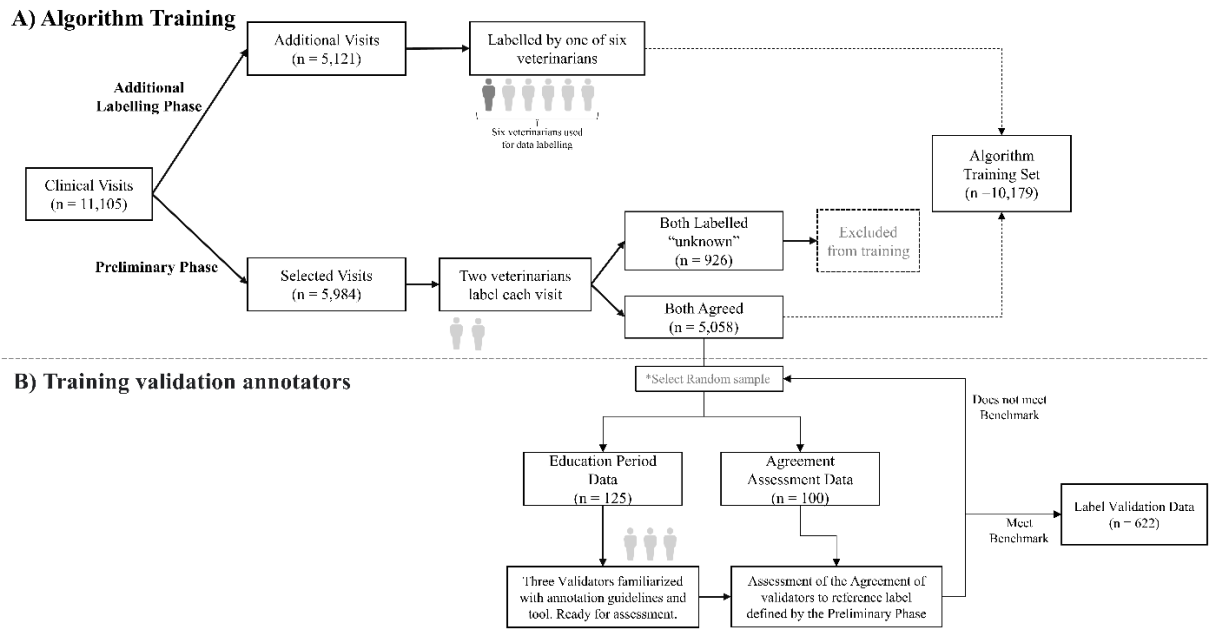

* Education period data was selected from a stratified random sample using 125 visits (25 from each of the five classes - as wellness, non-wellness, boarding, grooming, and retail). Agreement assessment data was a random sample selected to match the distribution of the five classes as seen in production.

**Supplemental Figure 2.** Calibration curve of actual versus predicted probability of the classification of wellness visits. The dashed black line represents the Youden equivalency line (slope=1.0, intercept=0.0), the solid red line represents the bias-corrected fit. The bootstrapped model was run with 2000 repetitions with a sample of 622 and gave a mean absolute error of 0.012.

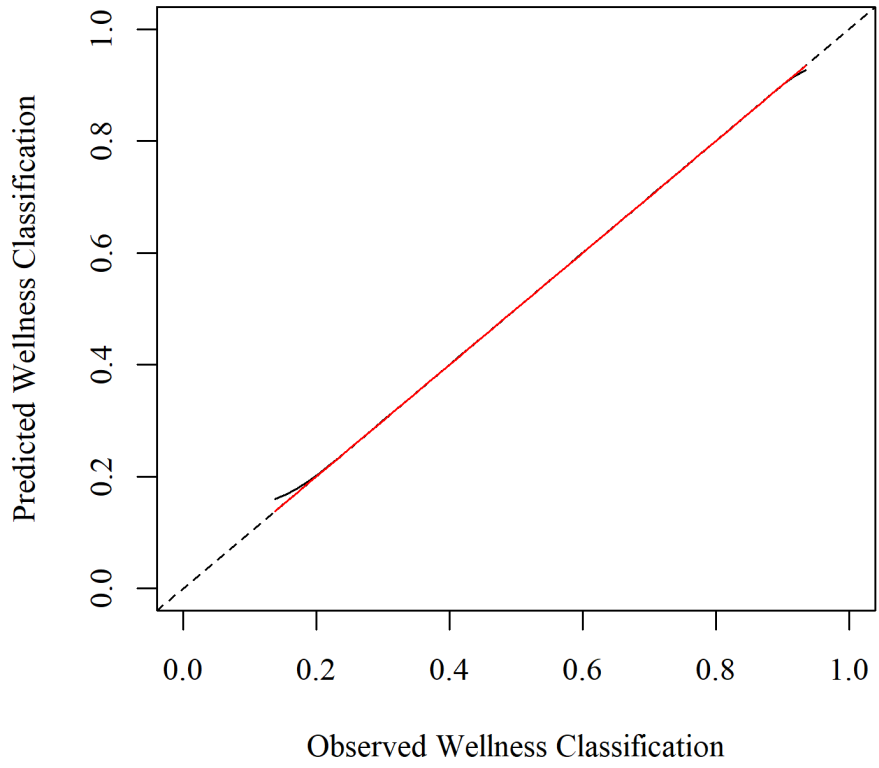